\documentclass[conference]{IEEEtran}
\IEEEoverridecommandlockouts
\usepackage{cite}
\usepackage{amsmath,amssymb,amsfonts}
\usepackage{algorithmic}
\usepackage[dvipdfmx]{graphicx}
\usepackage{textcomp}
\usepackage{xcolor}
\usepackage{bm}
\def\BibTeX{{\rm B\kern-.05em{\sc i\kern-.025em b}\kern-.08em
    T\kern-.1667em\lower.7ex\hbox{E}\kern-.125emX}}

\newcommand{\wfig}[1]{Fig.~\ref{fig:#1}}%
\newcommand{\wfigure}[1]{Figure~\ref{fig:#1}}%
\newcommand{\wtab}[1]{Tab.~\ref{tab:#1}}%
\newcommand{\wtable}[1]{Table~\ref{tab:#1}}%
\newcommand{\weq}[1]{Eq. (\ref{eq:#1})}%

\newcommand{\bx}{\bm{x}}
\newcommand{\bX}{\bm{X}}
\newcommand{\bc}{\bm{c}}
\newcommand{\xh}{\bm{x}_H}

\newcommand{\ch}{\bm{c}_H}

\newcommand{\bXL}{\bX_L}
\newcommand{\bXdL}{\bX'_L}

\begin{document}

\title{JPEG XT Image Compression with Hue Compensation for Two-Layer  HDR Coding}

\author{\IEEEauthorblockN{Hiroyuki KOBAYASHI}
\IEEEauthorblockA{
\textit{Tokyo Metropolitan College of Industrial Technology},\\
 Tokyo, Japan}
\and
\IEEEauthorblockN{Hitoshi KIYA}
\IEEEauthorblockA{
\textit{Tokyo Metropolitan University},\\
 Tokyo, Japan}
}

\maketitle

\begin{abstract}
We propose a novel JPEG XT image compression with hue compensation for two-layer HDR coding.
LDR images produced from JPEG XT bitstreams have some distortion in hue due to tone mapping operations.
In order to suppress the color distortion, we apply a novel hue compensation method based on the maximally saturated colors.
Moreover, the bitstreams generated by using the proposed method are fully compatible with the JPEG XT standard.
In an experiment, the proposed method is demonstrated not only to produce images with small hue degradation but also to maintain well-mapped luminance, in terms of three kinds of criterion: TMQI, hue value in CIEDE2000, and the maximally saturated color on the constant-hue plane.
\end{abstract}


\section{Introduction}
The interest of high dynamic range (HDR) imaging has recently been increasing in various area: photography, medical imaging, computer graphics, on vehicle cameras, astronautics.
HDR images have the information of the wide dynamic range of real scenes.
However, commonly used display devices which can directly represent HDR images are not popular yet.
Therefore, various two-layer encoding method to generate bitstreams with base layer and residual layer, have been proposed for compressing HDR images \cite{4517823,6287996,6411962,6637869,8456254, 8351220, 7991151}.
ISO/IEC JTC 1/SC 29/WG 1 (JPEG) has developed a series of international standards referred to as the JPEG XT\cite{6737677,Artusi2015,7426553,JPEGXT,7535096}.
The JPEG XT has been designed to be backward compatible with the legacy JPEG standard with two-layer coding.

For the base-layer coding, a low dynamic range (LDR) image is generated by a tone-mapping operation (TMO).
In the operation, the range of the luminance of an HDR image is compressed by using a tone-mapping (TM) operator, and then an LDR image is produced by combining the compressed luminance and the color information of the original HDR image.
However, TMOs only focusing on the luminance cause image colors to be distorted as pointed out in \cite{doi:10.1111/j.1467-8659.2009.01358.x, pureColorPreserving, 6336819, 1257395, 4378954}.

In this paper, we propose a novel JPEG XT image compression with hue compensation for two-Layer HDR coding.
The proposed compensation method utilizes the hue information based on the maximally saturated colors\cite{8451308}, for suppressing color distortions due to the influence of TMOs.

To evaluate the effectiveness of the proposed method, we perform a number of simulations.
In the simulations, the proposed method is compared with conventional TMOs in terms of the hue difference used in CIEDE2000\cite{doi:10.1002/col.1049}, the maximally saturated color difference, and Tone Mapped image Quality Index (TMQI)\cite{6319406}.
\section{Proposed Method}

\subsection{Overview}
\begin{figure}[tbp]
 \centering
 \includegraphics[width=8cm]{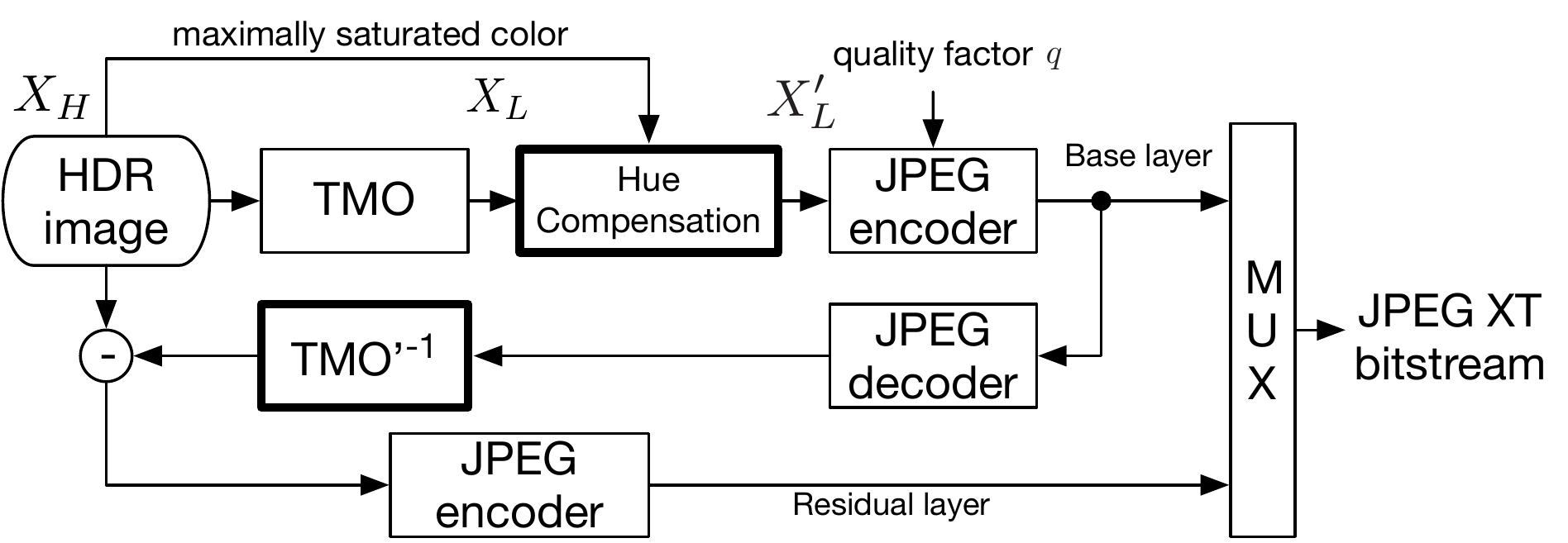}
 \caption{Block diagram of proposed two-layer coding. Thick frame boxes are modification parts from a JPEG XT encoder.}
 \label{fig:proposed}
\end{figure}

As shown in \wfig{proposed}, a JPEG XT bitstream consists of base layer and residual layer.
A tone-mapped LDR image is decoded from the base layer, and an HDR image is decoded by using both layers.
However, the LDR image includes some hue distortion due to the influence of TMO.
In order to suppress the hue distortion, we apply a hue compensation process based on the constant hue plane in the RGB color space before encoding LDR images.
In the hue compensation process, a hue compensated image $\bXdL$ is generated by using an LDR image $\bXL$ and the maximally saturated colors of the original HDR image.
In addition, inverse TMO (TMO$'^{-1}$ in \wfig{proposed}) is modified according to the hue compensation.
These modifications are applied to only the JPEG XT encoder, and moreover generated bitstreams are fully compatible with the JPEG XT standard.
As a result, LDR and HDR images are decoded from the bitstreams by using the standard JPEG XT decoder without any modification.

\subsection{Constant hue plane}
We focus on the constant hue plane in the RGB color space\cite{8451308}.
An input image $X_L$ is a 24-bit full color LDR image and each pixel of the image is represented as $\bx \in [0, 1]^3$.
$x_r$, $x_g$, and $x_b$ are the R, G, and B components of a pixel $\bx$, respectively.
In the RGB color space, a set of pixels which has the same hue forms a plane, called constant hue plane.
The shape of the constant hue plane is the triangle whose vertices correspond to white ($\bm{w} = (1, 1, 1)$), black ($\bm{k} = (0, 0, 0)$) and the maximally saturated color ($\bc$) with the same hue as $\bx$.
On the constant hue plane, a pixel $\bx = x_r\bm{r} + x_g\bm{g} + x_b\bm{b}$ can be represented as a linear combination as
\begin{equation}
\bx = a_w\bm{w} + a_k\bm{k} + a_c\bc,
\label{eq:x}
\end{equation}
where
\begin{equation}
\left\{\begin{array}{l}
a_w = \min(\bx)\\
a_k = \max(\bx) - \min(\bx)\\
a_c = 1 - \max(\bx)\\
\end{array}\right.,
\label{eq:awakac}
\end{equation}
and
$\max(\cdot)$ and $\min(\cdot)$ are functions that return the maximum and minimum elements of the pixel $\bx$, respectively.
Since $\bm w$, $\bm k$, $\bm c$ and $\bx$ exist on the plane and $\bx$ is an interior point of $\bm w$, $\bm k$ and $\bm c$, the following equations hold.
\begin{align}
&a_w + a_k + a_c = 1,\\
&0\leq a_w, a_k, a_c \leq 1.
\end{align}

Moreover, the maximally saturated color $\bc = (c_r, c_g, c_b)$ is given by
\begin{equation}
\left\{\begin{array}{l}
c_r = (x_r - \min(\bx))/(\max(\bx)-\min(\bx))\\
c_g = (x_g - \min(\bx))/(\max(\bx)-\min(\bx))\\
c_b = (x_b - \min(\bx))/(\max(\bx)-\min(\bx))
\end{array}\right..
\label{eq:msc}
\end{equation}
From \weq{msc}, the elements of $\bm c$ corresponding to the maximum and minimum elements of the pixel $\bx$ become 1 and 0, respectively.

We consider the same pixel representation on the constant hue plane for HDR images and discuss the relation between HDR images and LDR ones.
Each pixel of an HDR image $X_H$ is represented as $\xh = (x_{Hr}, x_{Hg}, x_{Hb})$ where $x_{Hr}$, $x_{Hg}$, and $x_{Hb}$ are generally real numbers.
As well as LDR images, we define white, black and the maximally saturated color as $\bm{w}=(1, 1, 1)$, $\bm{k}=(0, 0, 0)$ and $\ch$, 
where $\ch$ is calculated from \weq{msc} by replacing $\bx$ with $\xh$, and then the pixel value $\xh$ is also represented as a linear combination as
\begin{equation}
\xh = a_{Hw}\bm{w} + a_{Hk}\bm{k} + a_{Hc}\ch,
\label{eq:xh}
\end{equation}
where $a_{Hw}$, $a_{Hk}$ and $a_{Hc}$ are coefficients calculated from $\xh$, and the maximally saturated color $\ch = (c_{Hr}, c_{Hg}, c_{Hb})$ is given by
\begin{equation}
\left\{\begin{array}{l}
c_{Hr} = (x_{Hr} - \min(\xh))/(\max(\xh)-\min(\xh))\\
c_{Hg} = (x_{Hg} - \min(\xh))/(\max(\xh)-\min(\xh))\\
c_{Hb} = (x_{Hb} - \min(\xh))/(\max(\xh)-\min(\xh))
\end{array}\right..
\label{eq:msch}
\end{equation}
.

In general, $\ch = \bm{c}$  is not satisfied, because some hue distortion occurs due to the influence of TMO.
Therefore, to correct the hue distortion, we replace $\bx$ with $\bx'$ as,
\begin{align}
\bx' &= a_{w}\bm{w} + a_{k}\bm{k} + a_{c}\ch\nonumber\\
	&= x'_r\bm{r} + x'_g\bm{g} + x'_b\bm{b},
	\label{eq:bxd}
\end{align}
where 
\begin{equation}
\left\{\begin{array}{l}
x'_r = a_w + a_c\cdot c_{Hr}\\
x'_g = a_w + a_c\cdot c_{Hg}\\
x'_b = a_w + a_c\cdot c_{Hb}\\
\end{array}\right..
\end{equation}
This process corresponds to `Hue Compensation' in \wfig{proposed}.
Note that $\bx'$ and $\xh$ are on the same constant hue plane.

\subsection{Procedure of hue compensation}
The procedure of the proposed method shown in \wfig{proposed} is shown as follows.
\begin{enumerate}
\item Generate an LDR image $X_L$ by using a TMO.
\item Calculate $a_w$, $a_k$, and $a_c$ for each pixel value $x$ of $X_L$ by \weq{awakac}.
\item Calculate $\ch$ for each pixel value $x_H$ of the original HDR image $X_H$ according to \weq{msch}.
\item Generate a compensated image $X'_L$ by \weq{bxd}.
\item Input $X'_L$ to a JPEG XT encoder.
\end{enumerate}

\begin{figure}[tbp]
 \centering
 \includegraphics[width=6cm]{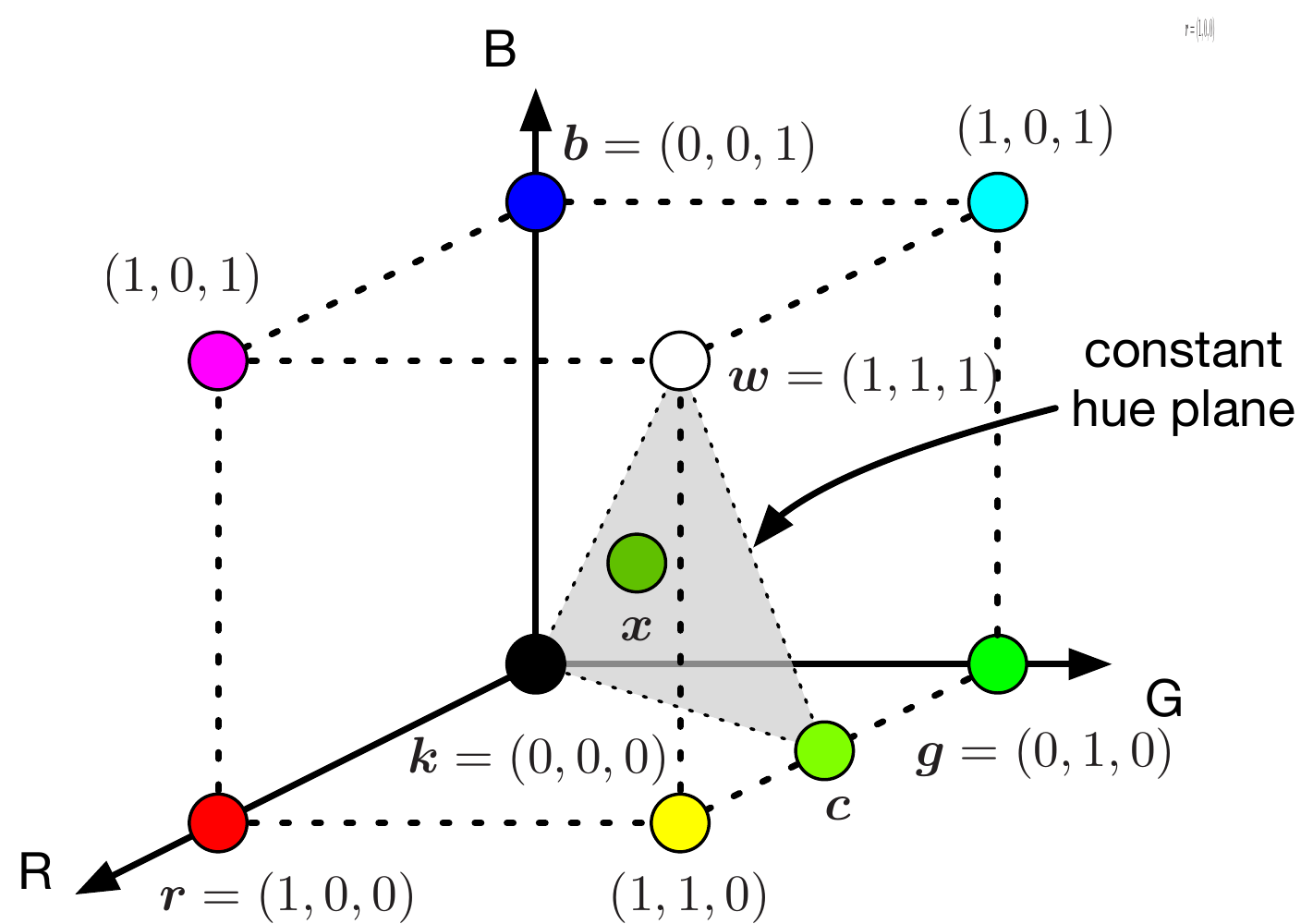}
 \caption{Constant hue plane in the RGB color space}
 \label{fig:colorSpace}
\end{figure}

\section{Experimental results}

\subsection{Hue distortion}
We used five HDR images selected from an HDR image database\cite{fairchild2007hdr}.
We adapted two objective metrics to evaluate hue distortion caused in TMOs under the use of three TMOs: JPEG XT default TMO, Reinhard's local operator\cite{Reinhard2002}, and Drago's TMO\cite{drago2003adaptive}.
One is $\Delta c$, which is the average of euclidean norms of the maximally saturated colors $\ch$ and $\bm c$.
The other is the hue differences defined in CIEDE2000 $\Delta H$, which was published by the CIE\cite{doi:10.1002/col.1049}.

\wtable{HueDistortion} shows evaluation results in terms of the objective metrics.
In the table, the proposed method outperformed the conventional JPEG XT bitstreams for both TMOs.
Therefore, the proposed method is demonstrated to be effective for improving hue distortion in terms of not only $\Delta c$ but also $\Delta H$.

\begin{table}[tbp]
\centering
\caption{Hue distortion (quality factor $q=80$)}
\label{tab:HueDistortion}
(a) JPEG XT default\\
\begin{tabular}{l|c@{ }c|c@{ }c}
&
\multicolumn{2}{c|}{$\Delta c$} &
\multicolumn{2}{c}{$\Delta H$}\\
Images & Conventional & Proposed & Conventional & Proposed \\ \hline
McKeesPub & 
0.135793	& \textbf{0.059346} &
8.909750 & \textbf{4.468197}\\
BloomingGorse2 &
0.144254	&	\textbf{0.063077} &
9.580835	&	\textbf{3.446832} \\
WillyDesk &
0.093953	&	\textbf{0.082277} &
1.576254	&	\textbf{0.859007}\\
CanadianFalls &
0.152728	&	\textbf{0.127771} &
1.964077	&	\textbf{1.388976}\\
YosemiteFalls &
0.127535	&	\textbf{0.105994} &
4.382326	&	\textbf{2.791083}\\
\end{tabular}
(b) Reinhard's local\\
\begin{tabular}{l|c@{ }c|c@{ }c}
&
\multicolumn{2}{c|}{$\Delta c$} &
\multicolumn{2}{c}{$\Delta H$}\\
Images & Conventional & Proposed & Conventional & Proposed \\ \hline
McKeesPub & 0.135598	& \textbf{0.045975} &
14.294454 & \textbf{4.775688}\\
BloomingGorse2 &
0.093910	&	\textbf{0.043285} &
8.098938 &	\textbf{3.313652}\\
WillyDesk &
0.081648 &	\textbf{0.057996} &
2.64185 &	\textbf{0.899966}\\
CanadianFalls &
0.101548	&	\textbf{0.084915} &
1.686966	&	\textbf{1.243498}\\
YosemiteFalls &
0.095419	&	\textbf{0.077479} &
5.07198		&	\textbf{2.921161}\\
\end{tabular}\\
(c)Drago TMO\\
\begin{tabular}{l|c@{ }c|c@{}c}
&
\multicolumn{2}{c|}{$\Delta c$} &
\multicolumn{2}{c}{$\Delta H$}\\
Images & Conventional & Proposed & Conventional & Proposed \\ \hline
McKeesPub & 0.200437	& \textbf{0.052533} &
 16.642008 & \textbf{4.261362}\\
BloomingGorse2 &
0.255610	&	\textbf{0.156782} &
11.268738	&	\textbf{2.947961}\\
WillyDesk &
0.196852	&	\textbf{0.178788} &
1.486648	&	\textbf{0.889389}\\
CanadianFalls &
0.394086	&	\textbf{0.372989} &
1.784205	&	\textbf{1.155001}\\
YosemiteFalls &
0.290870	&	\textbf{0.265017} &
3.629349	&	\textbf{2.340995}\\
\end{tabular}
\end{table}

\subsection{Influence on tone mapping}
Next, we evaluate the quality of tone-mapped LDR images in terms of TMQI.
From \wtab{TMQI}, the proposed method is shown to offer almost the same TMQI scores as those of the conventional bitstreams.
The results show that the proposed one allows us to compensate hue distortion, while maintaining TM performances.

\begin{table}[tbp]
\centering
\caption{Tone mapping Image Quality Index (TMQI)}
\label{tab:TMQI}
(a) JPEG XT default\\
\begin{tabular}{l|c@{ }c}
Images & Conventional & Proposed \\ \hline
McKeesPub &
0.745649		& \textbf{0.748897} \\
BloomingGorse2 &
0.892965	&	\textbf{0.898172} \\
WillyDesk &
0.698159	&	\textbf{0.700518} \\
CanadianFalls &
0.849512	&	\textbf{0.853757} \\
YosemiteFalls &
0.895106	&	\textbf{0.896661} \\
\end{tabular}\\
(b) Reinhard's local\\
\begin{tabular}{l|c@{ }c}
Images & Conventional & Proposed \\ \hline
McKeesPub &
\textbf{0.786303} & 0.782583\\
BloomingGorse2 &
0.953723	&	\textbf{0.958084}\\
WillyDesk &
0.765819	&	\textbf{0.766119}\\
CanadianFalls &
0.935495	&	\textbf{0.937796}\\
YosemiteFalls &
\textbf{0.976044}	&	0.975148\\
\end{tabular}\\
(c) Drago TMO\\
\begin{tabular}{l|c@{ }c}
Images & Conventional & Proposed \\ \hline
McKeesPub &
0.787873		& \textbf{0.795471} \\
BloomingGorse2 &
0.818499	&	\textbf{0.843880} \\
WillyDesk &
0.712146	&	\textbf{0.717335} \\
CanadianFalls &
0.816237	&	\textbf{0.818759} \\
YosemiteFalls &
0.809861	&	\textbf{0.815588} \\
\end{tabular}

\end{table}

\subsection{Visual evaluation}
\begin{figure}[tbp]
\centering
\begin{tabular}{cc}
\includegraphics[height=25mm]{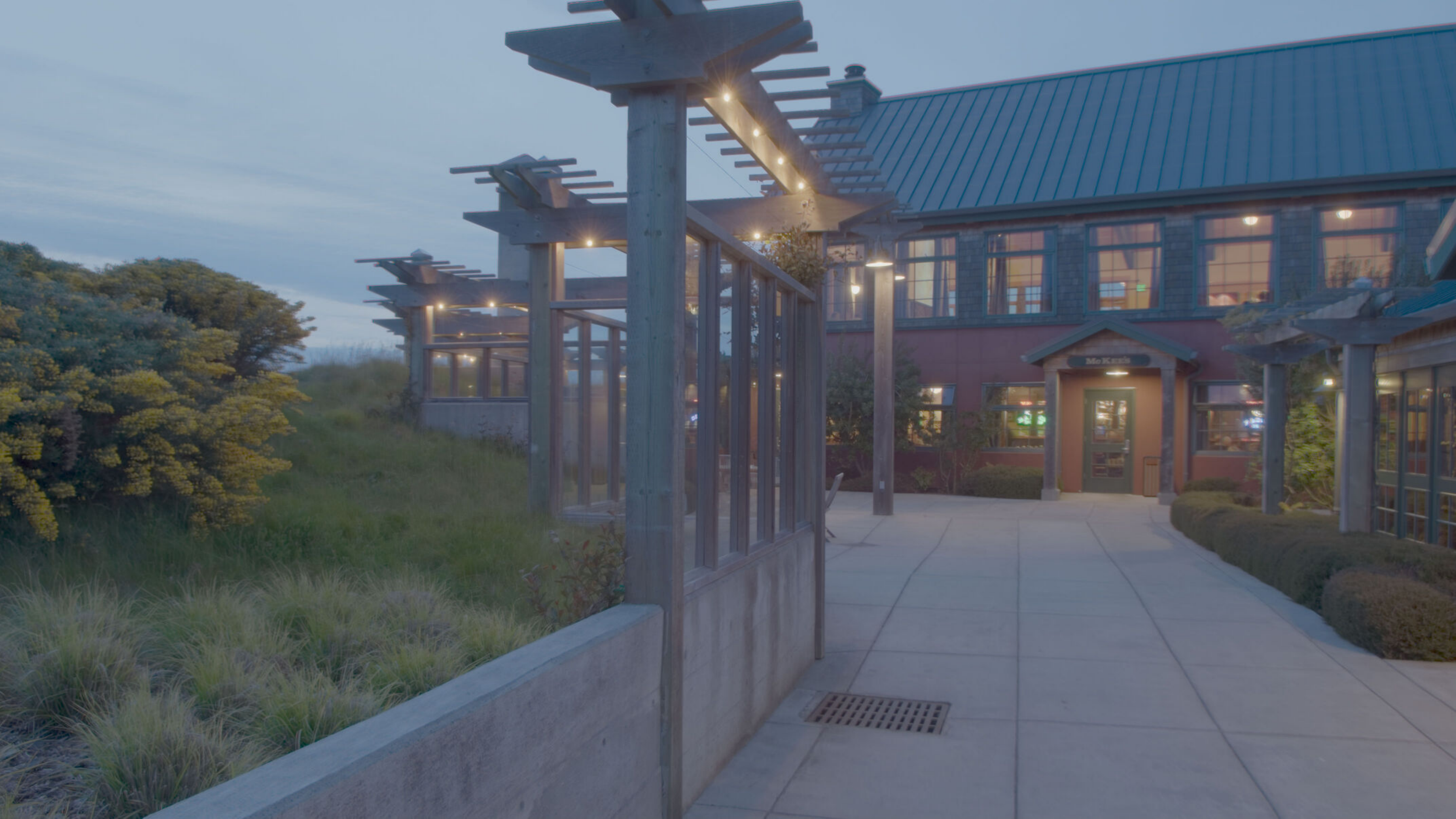}
&
\includegraphics[height=25mm]{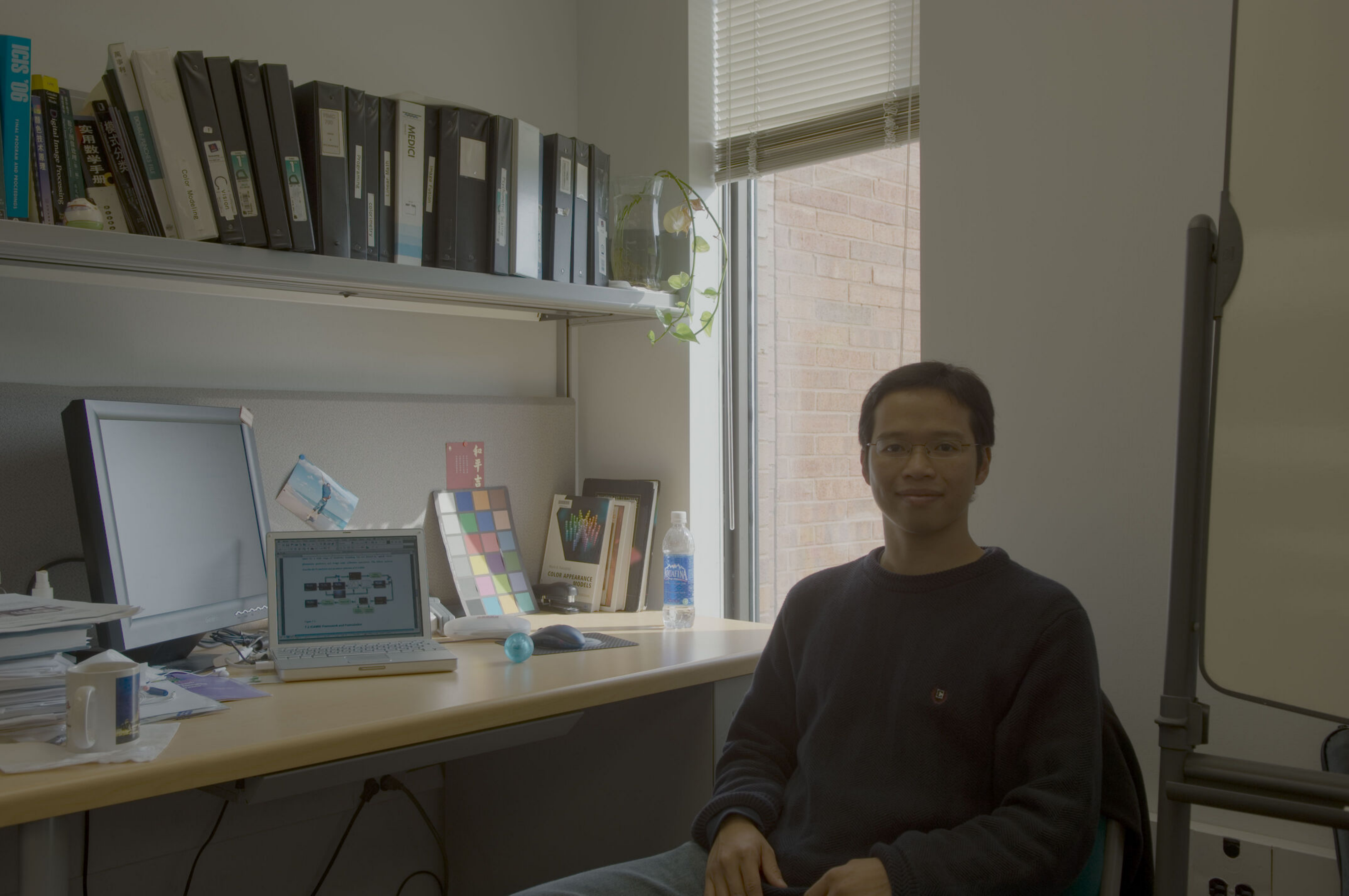}
\\
{\footnotesize (a) Conventional (McKeesPub)} &
{\footnotesize (b) Conventional (WillyDesk)}\\
\includegraphics[height=25mm]{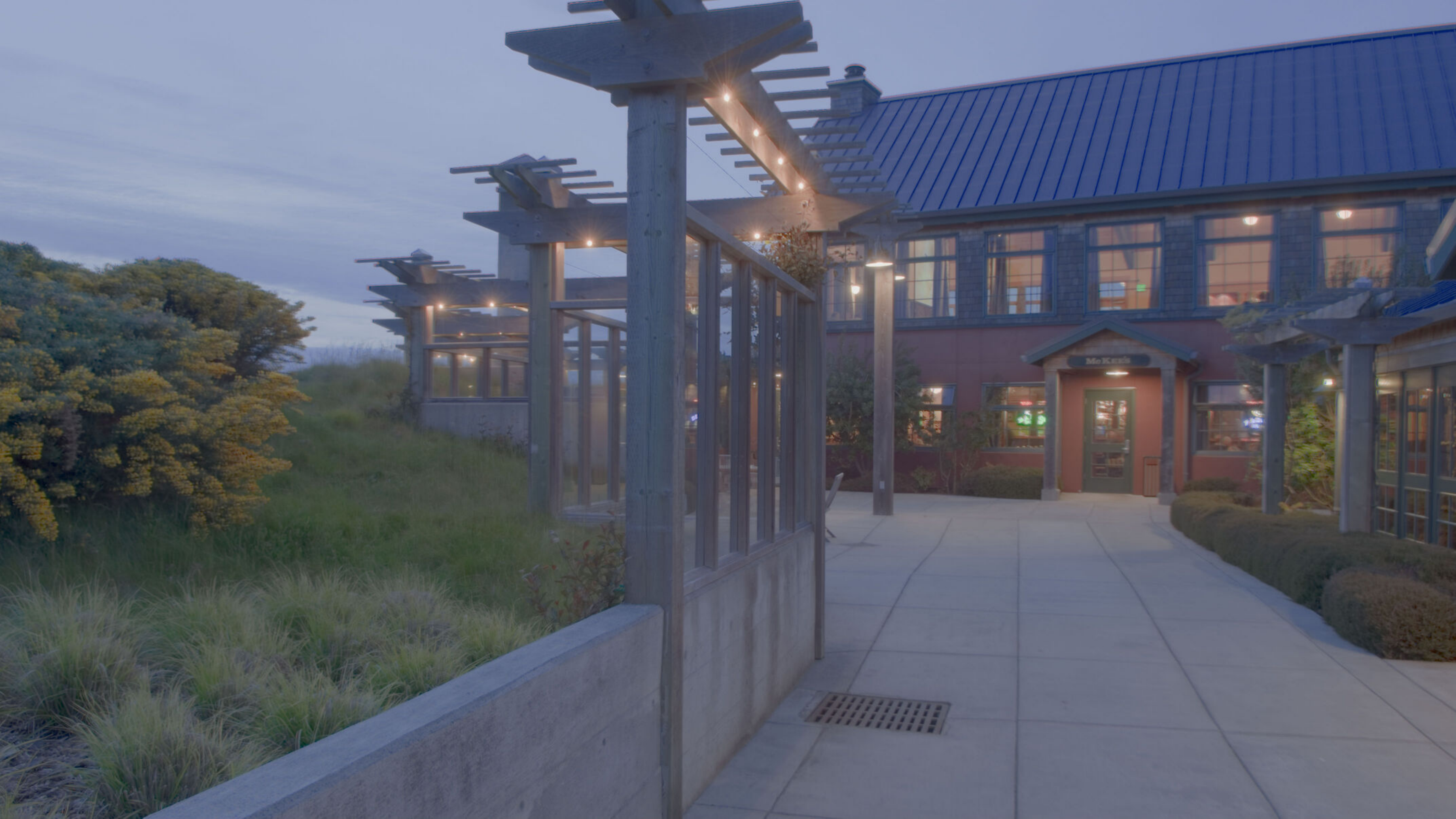}
&
\includegraphics[height=25mm]{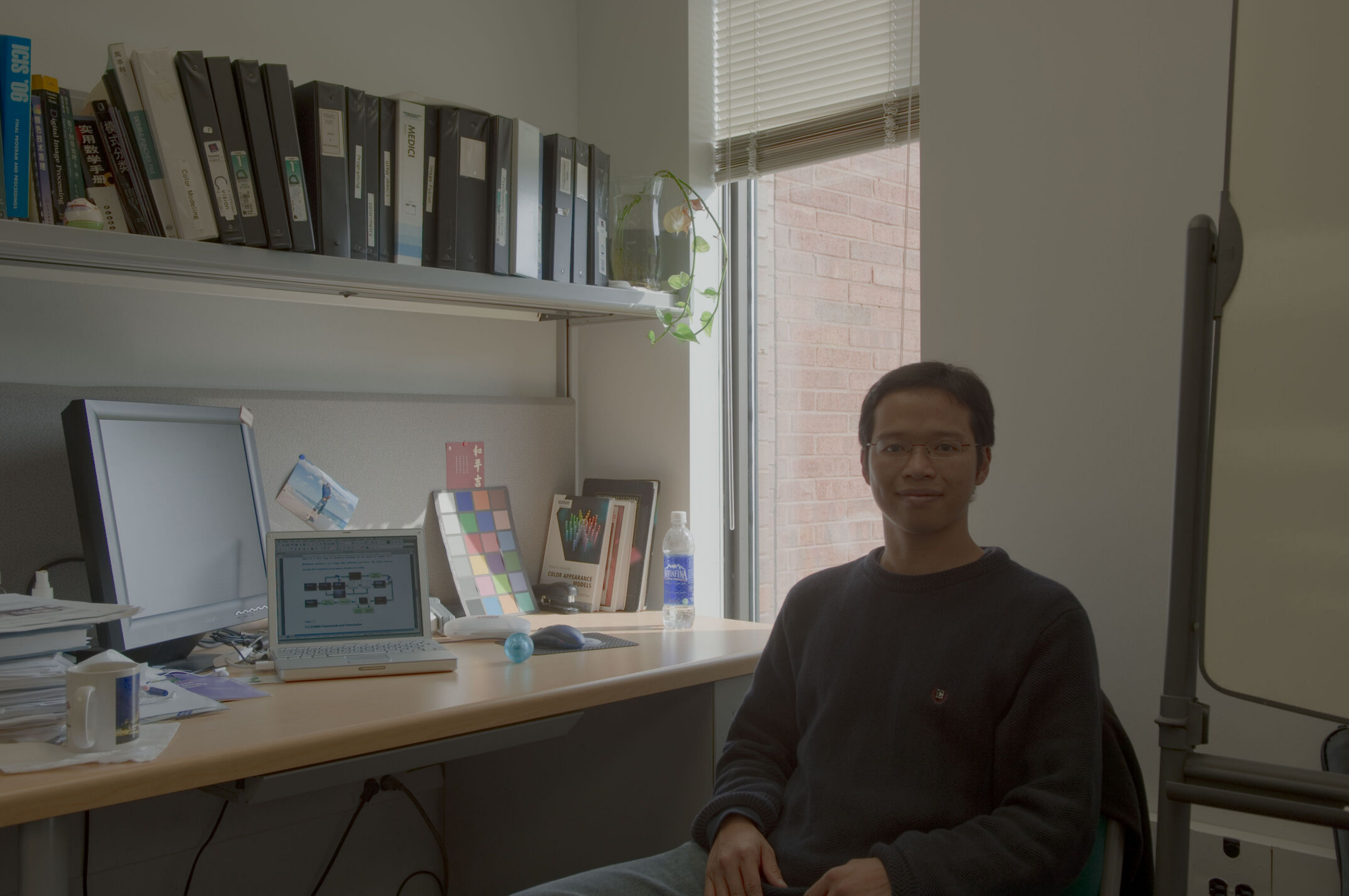}
\\
{\footnotesize (a) Proposed (McKeesPub)} &
{\footnotesize (b) Proposed (WillyDesk)}
\end{tabular}
\caption{Decoded LDR images}
\label{fig:org}
\end{figure}

We used two HDR images, `McKeesPub' and `WillyDesk', to visually evaluate the quality of images.
The images were encoded by the proposed encoder and the conventional JPEG XT one, respectively.
In this experiment, the default TM operator of JPEG XT was used as a TM operator.
\wfigure{org} shows decoded LDR images, where (a) and (b) were results produced by using the conventional approach, and (c) and (d) were generated by using the proposed one.
From the result, we can confirm there are color differences in some places.

\begin{figure}[tbp]
\centering
\begin{tabular}{cc}
\includegraphics[height=25mm]{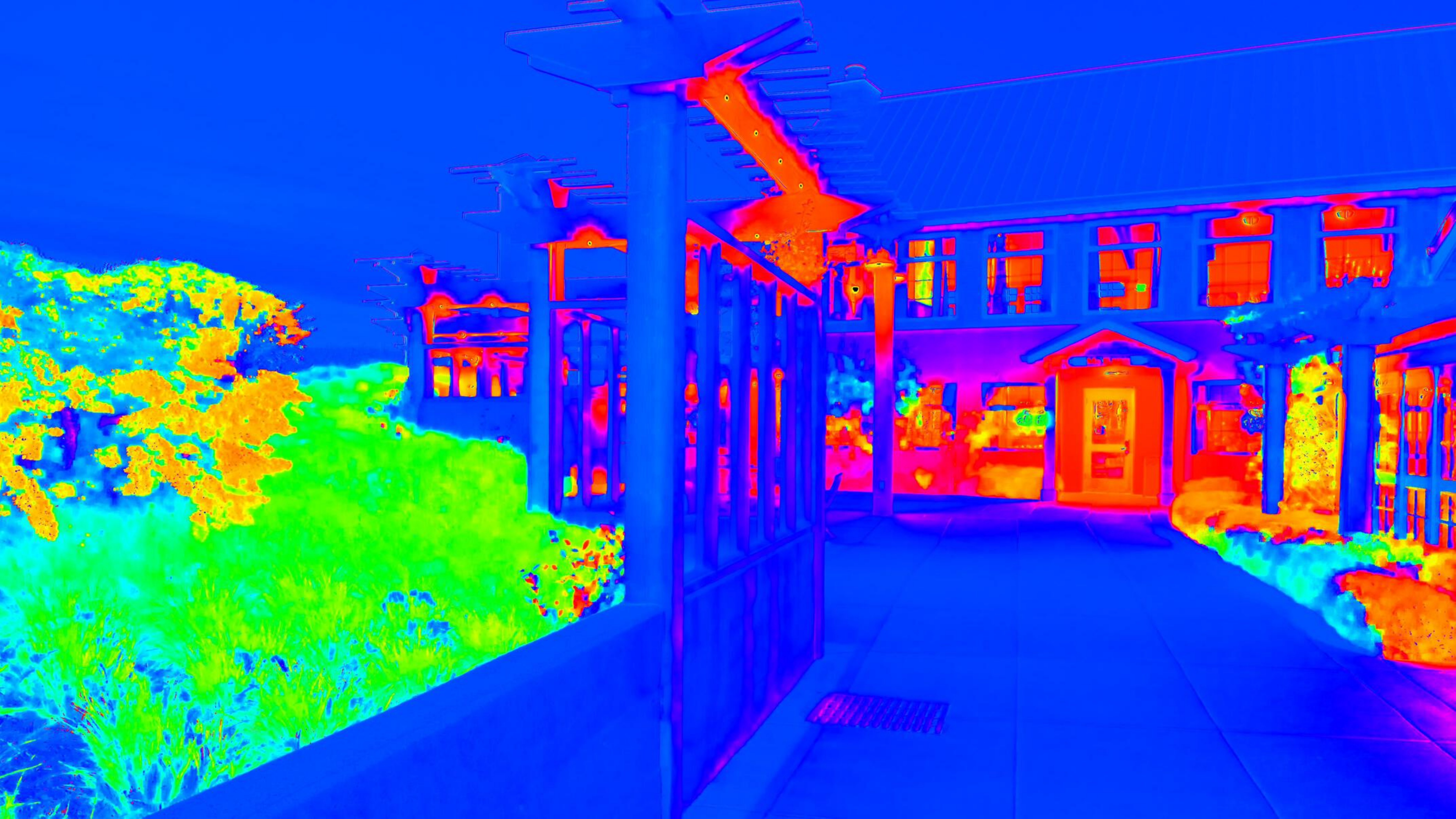} &
\includegraphics[height=25mm]{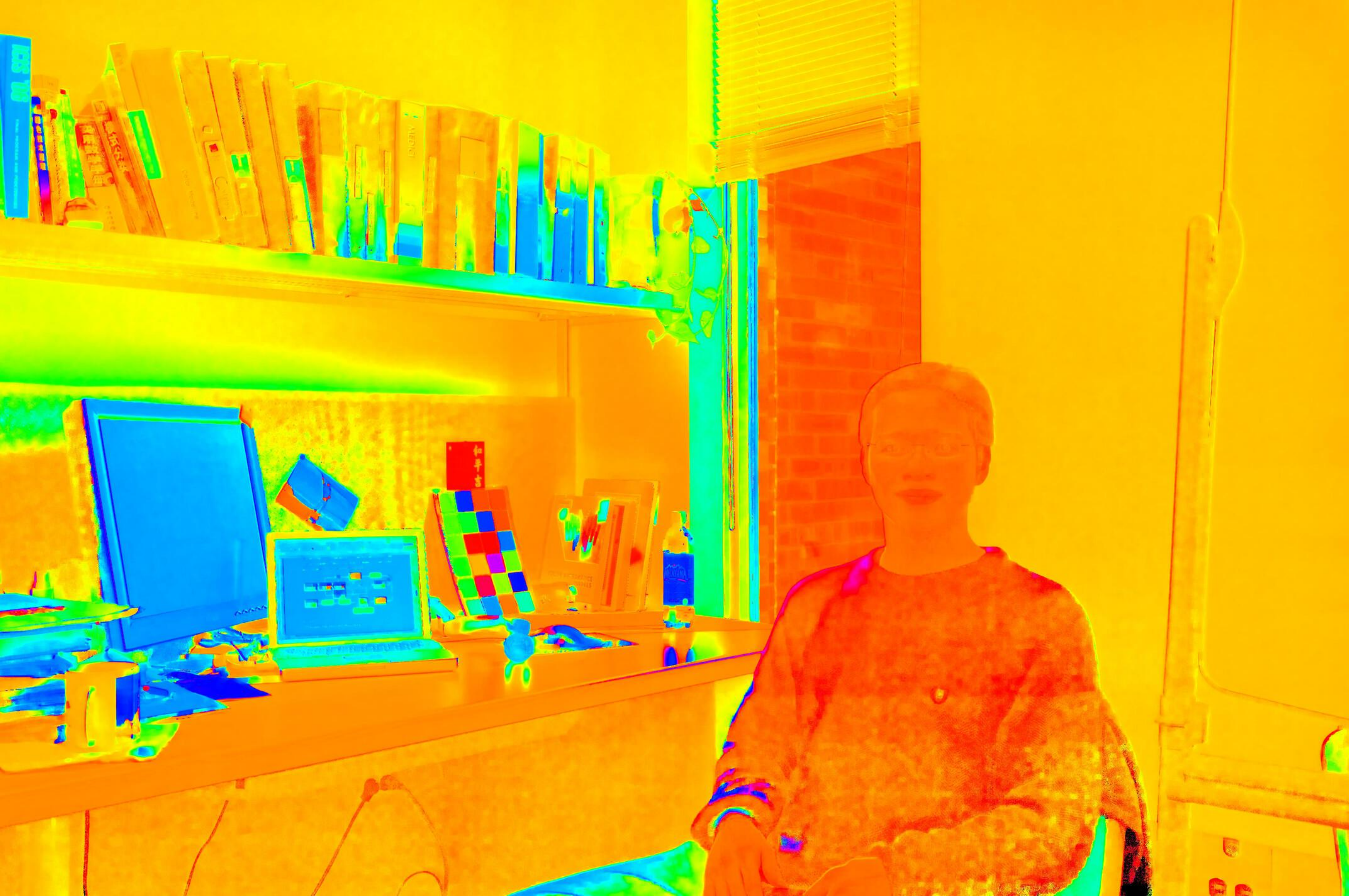}\\
{\footnotesize (a) HDR image (McKeesPub)} &
{\footnotesize (b) HDR image (WillyDesk)} \\
\includegraphics[height=25mm]{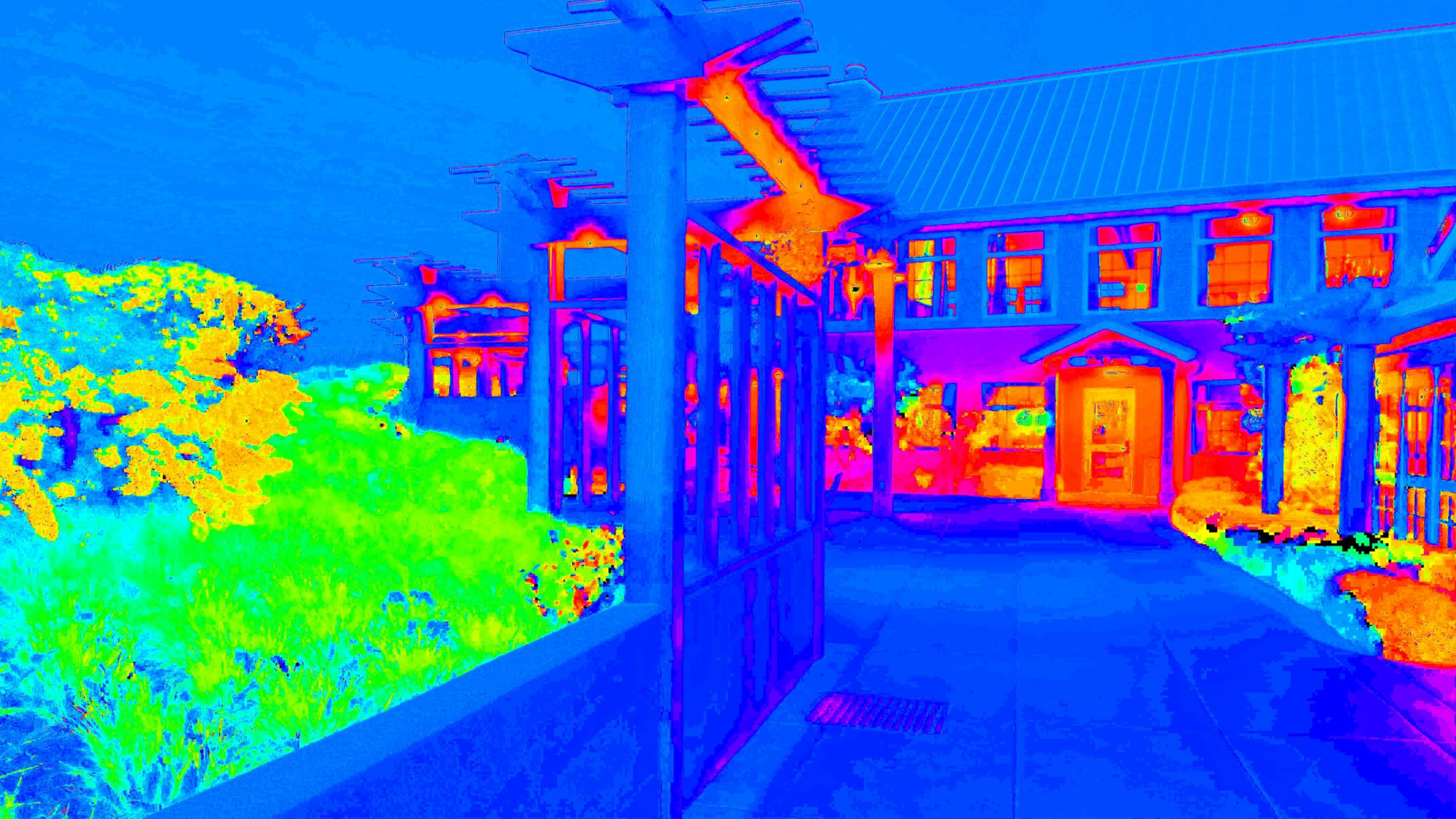} &
\includegraphics[height=25mm]{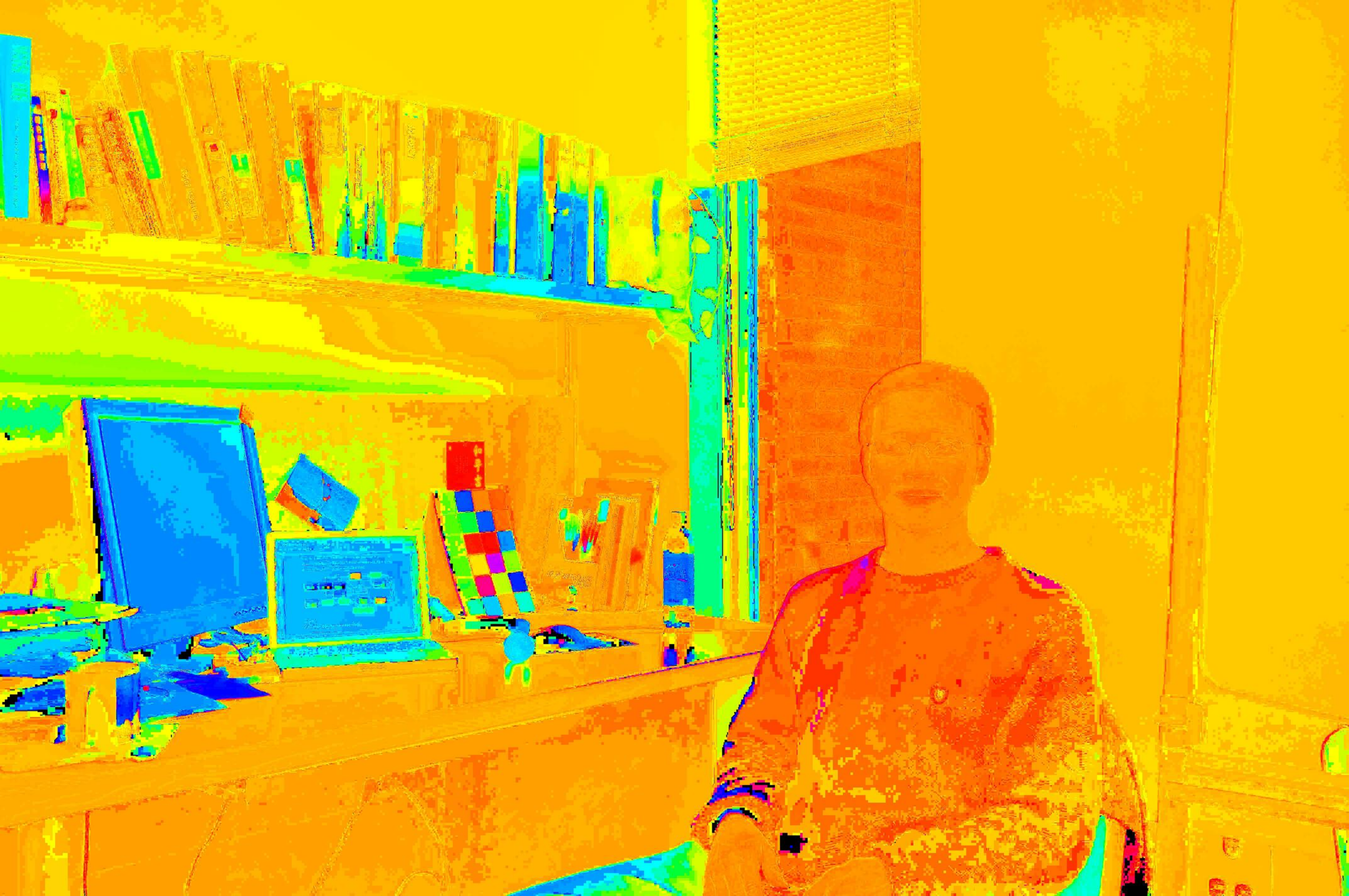} \\
{\footnotesize (c) Conventional (McKeesPub)} &
{\footnotesize (d) Conventional (WillyDesk)} \\
\includegraphics[height=25mm]{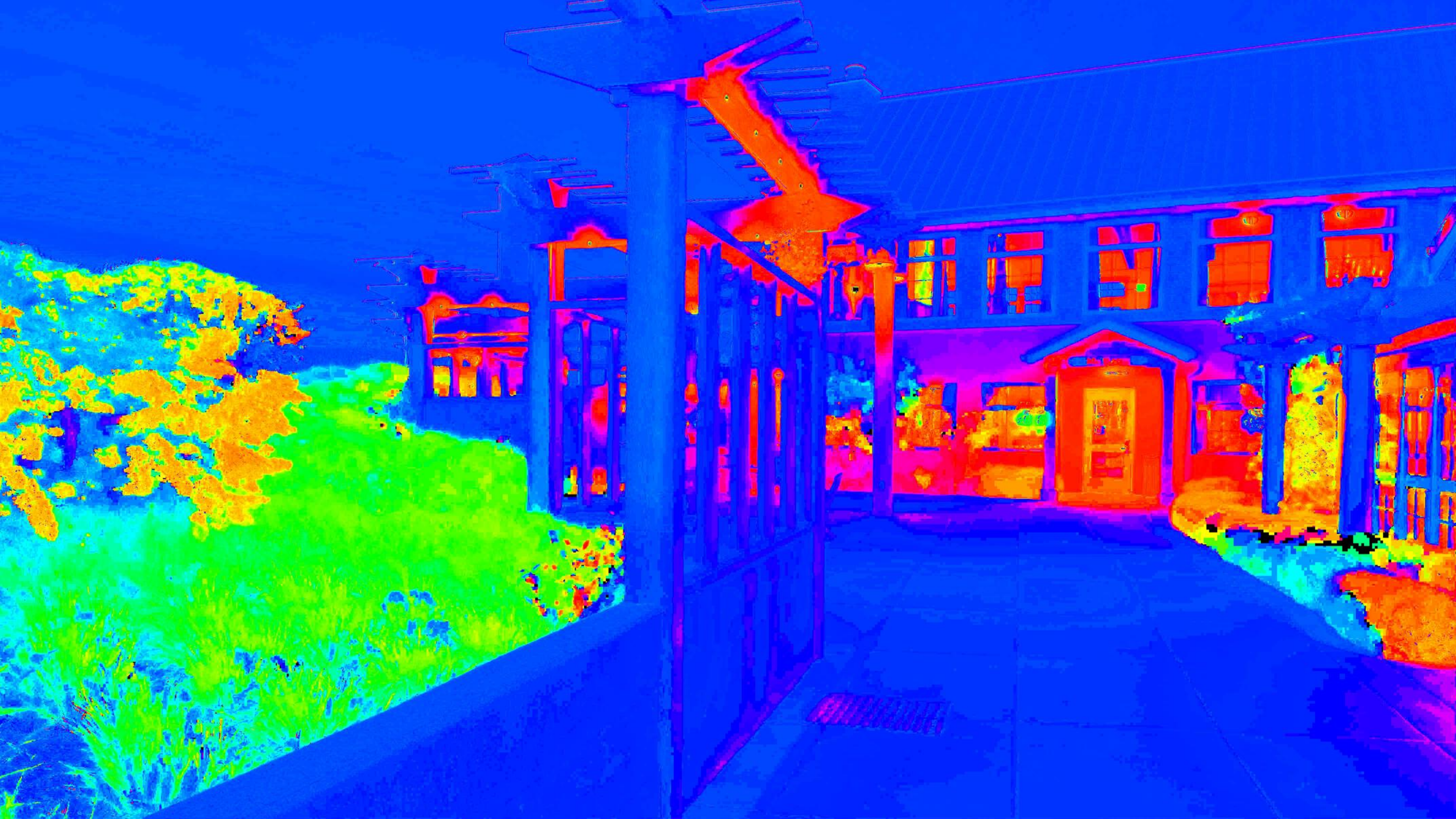} &
\includegraphics[height=25mm]{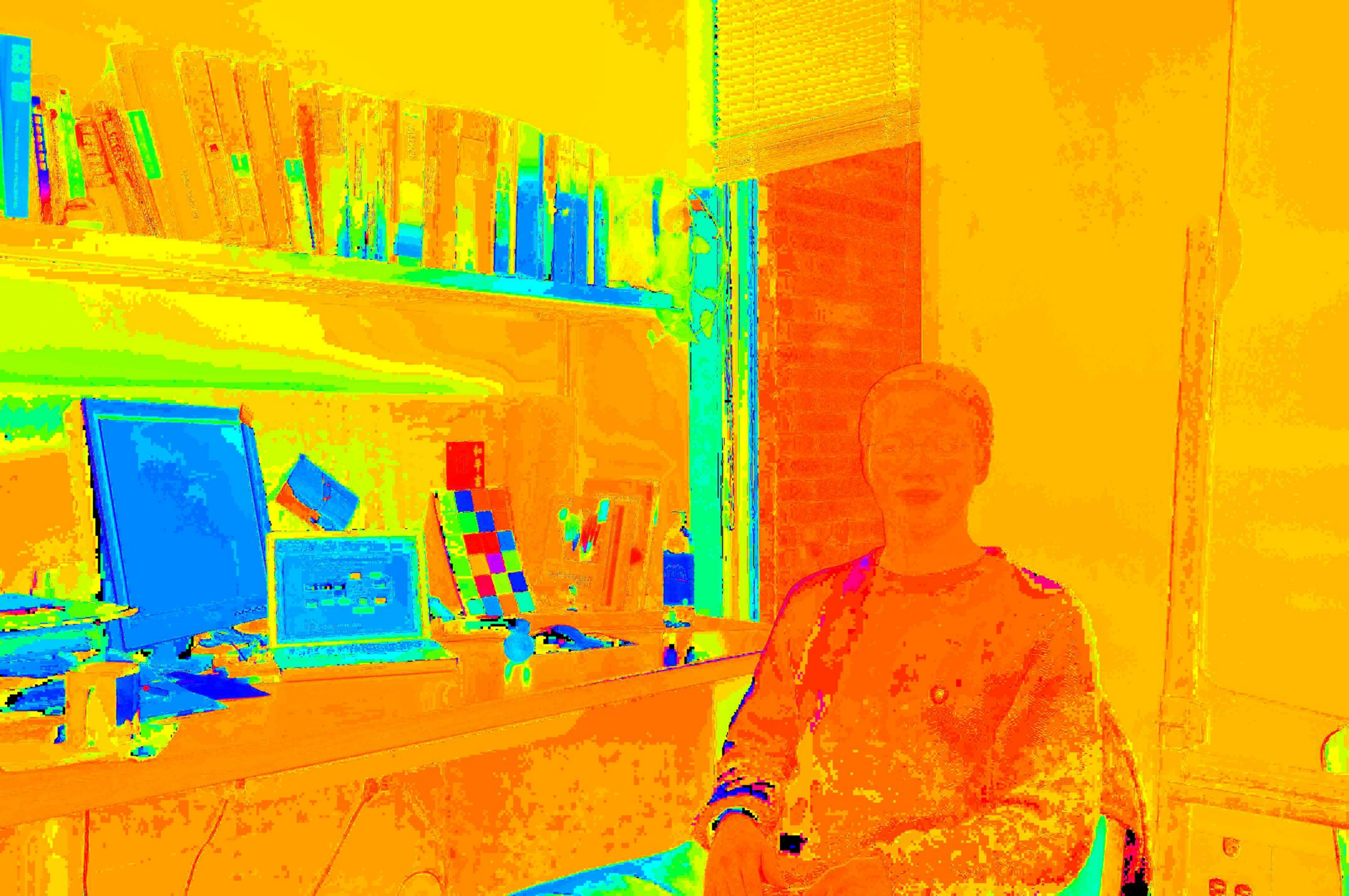} \\
{\footnotesize (e) Proposed (McKeesPub)} &
{\footnotesize (f) Proposed (WillyDesk)}\\
\end{tabular}
\caption{Maximally saturated color images generated from original HDR images and decoded LDR ones}
\label{fig:msc}
\end{figure}

To more clearly demonstrate the differences, maximally saturated color images are shown in \wfig{msc}.
The proposed method is demonstrated to have closer maximally saturated colors to those of the HDR images than the conventional method.



\begin{figure}[tbp]
\centering
\begin{tabular}{cc}
\includegraphics[height=20mm]{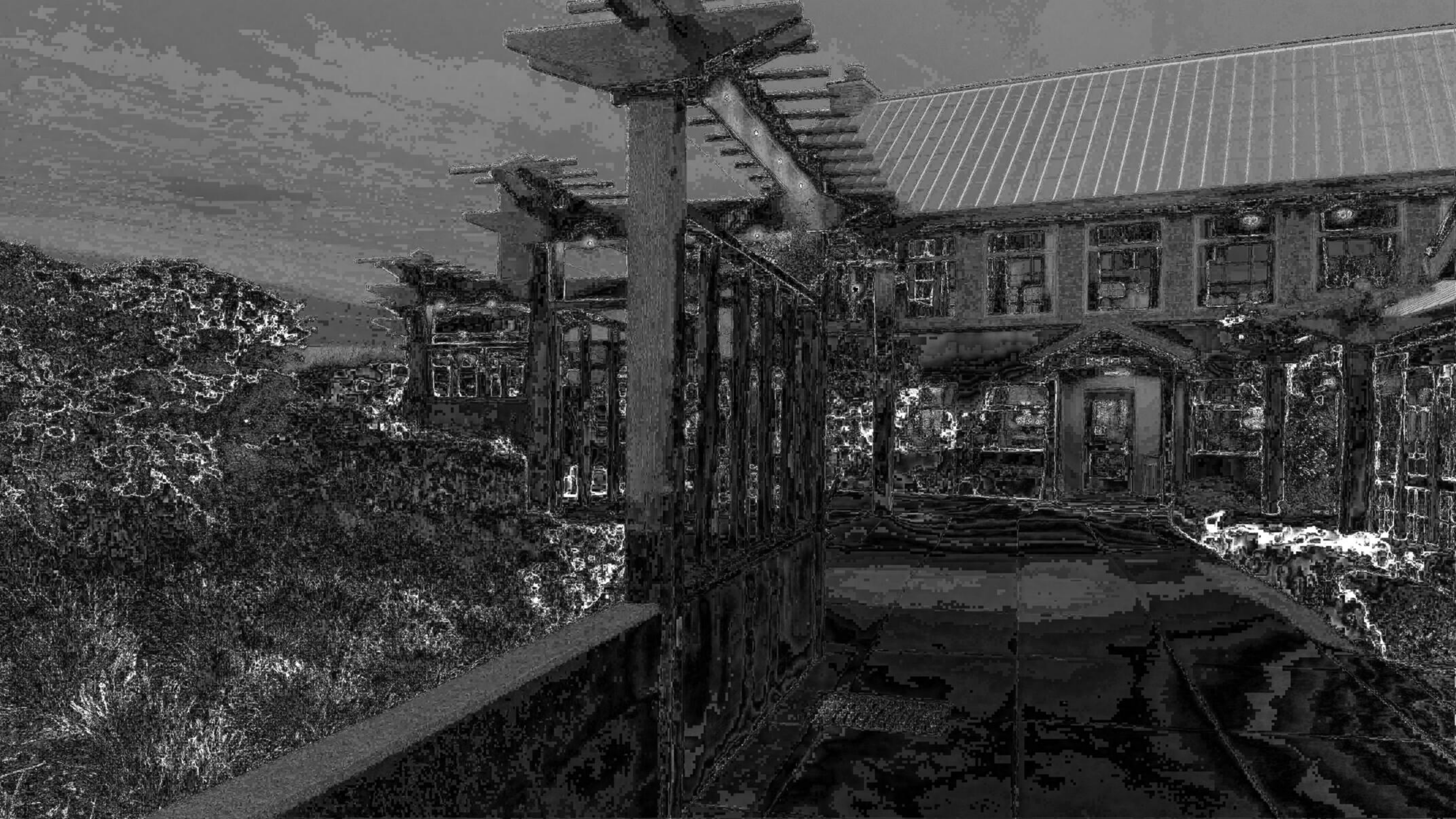}
\includegraphics[height=20mm]{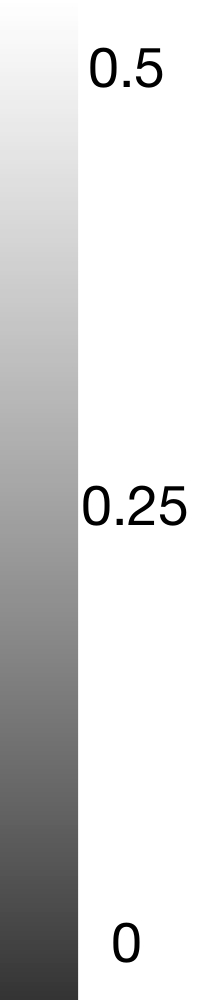}
&
\includegraphics[height=20mm]{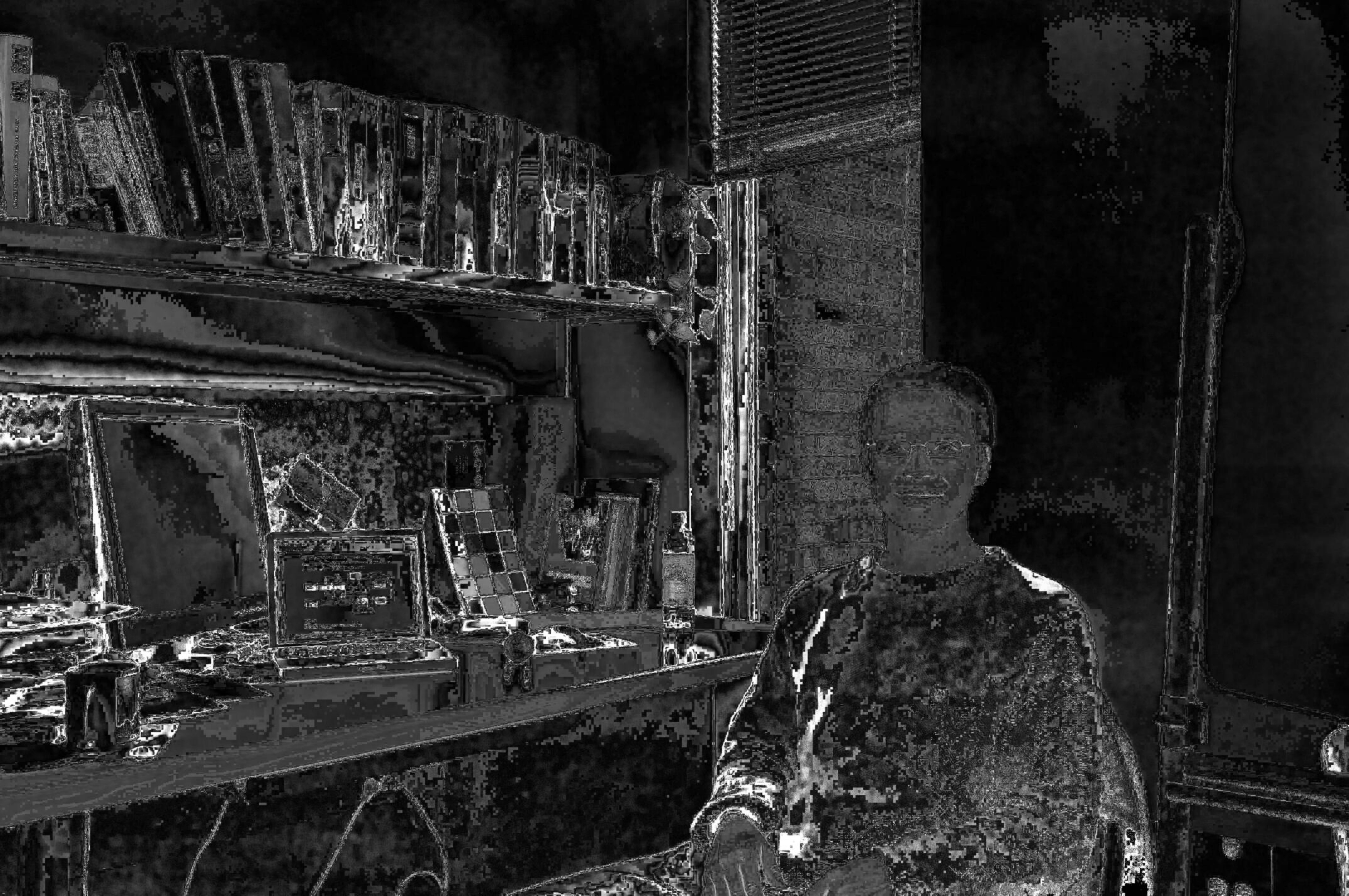}
\includegraphics[height=20mm]{colorBar.pdf}
\\
{\footnotesize (a) Conventional(McKeesPub)} & 
{\footnotesize (b) Conventional (WillyDesk)} \\
\includegraphics[height=20mm]{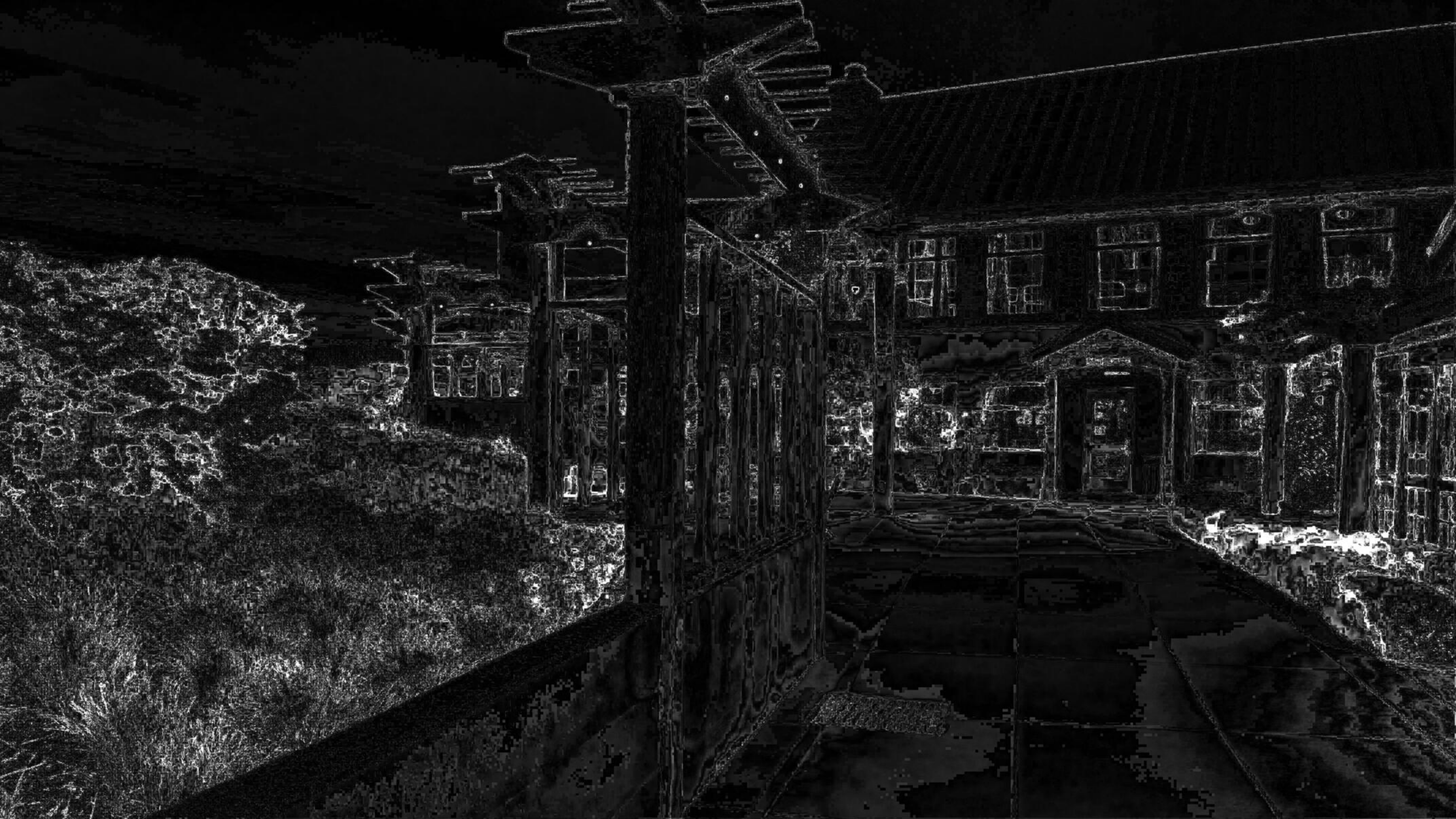}
\includegraphics[height=20mm]{colorBar.pdf}
&
\includegraphics[height=20mm]{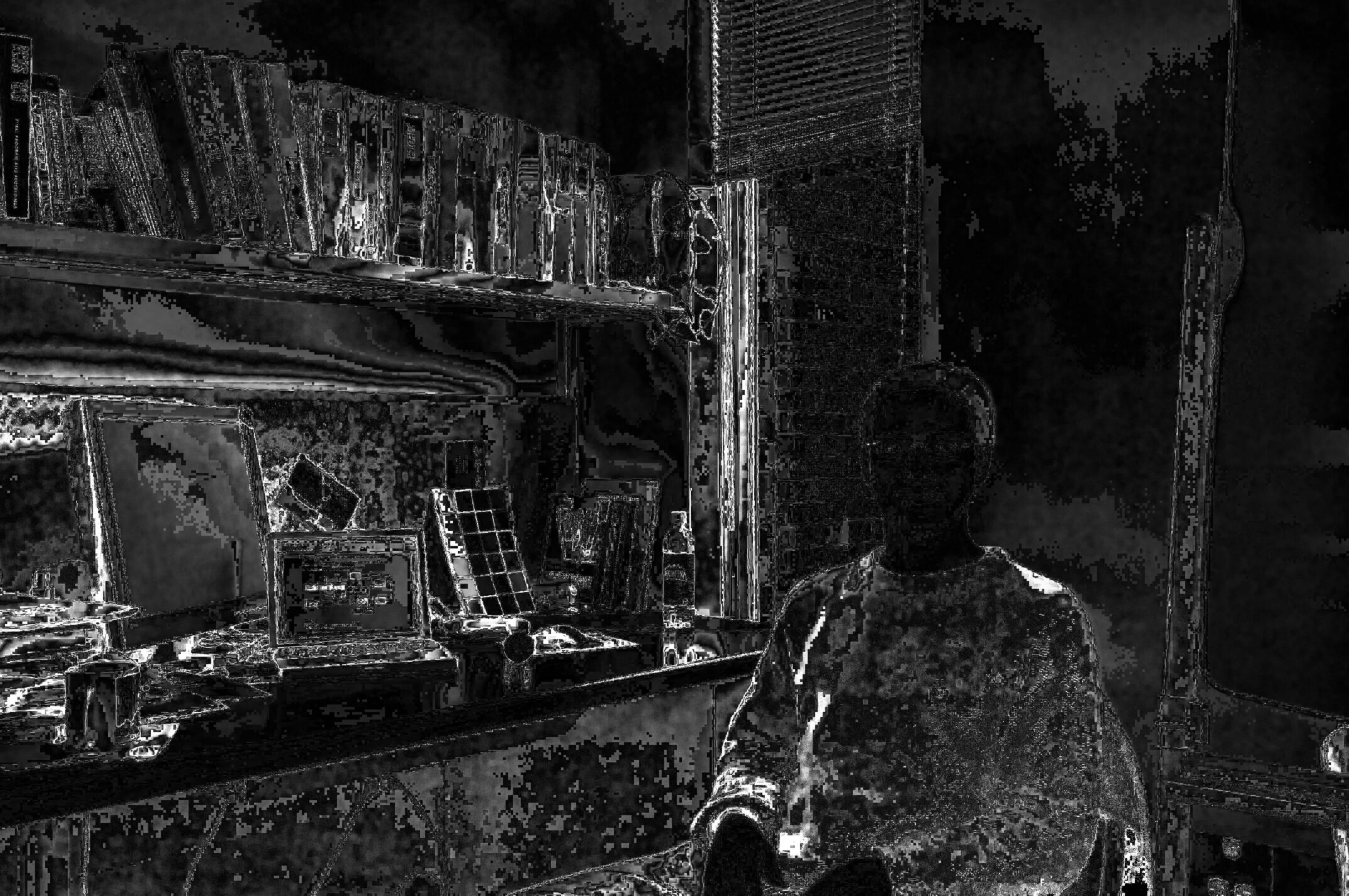}
\includegraphics[height=20mm]{colorBar.pdf}
\\
{\footnotesize (c) Proposed(McKeesPub)} &
{\footnotesize (d) Proposed (WillyDesk)}
\\
\end{tabular}
\caption{Difference of the maximally saturated color $\ch$ with $\bm c_D$ with JPEG XT default TMO}
\label{fig:difference}
\end{figure}

\wfigure{difference} shows euclidean norms of the maximally saturated color $\ch$ and $\bm c$.
From these figures, the proposed method is also demonstrated to be effective in reducing hue distortion.

\section{Conclusion}
In this paper, we proposed a novel two-layer HDR coding method to reduce hue distortion of LDR images due to the influence of TMO.
The proposed method enables us to compensate the hue distortion under the use of any TM operator, while maintaining well-mapped luminance.
In addition, the proposed one has compatibility with JPEG XT bitstreams.
Experimental results showed the effectiveness of the proposed method in terms of three objective metrics: $\Delta c$, $\Delta H$, and TMQI.

\bibliographystyle{IEEEbib}
\bibliography{./bibs/IEEEabrv,./bibs/refs}

\end{document}